\documentclass[10pt,twocolumn,letterpaper]{article}

 \usepackage[pagenumbers]{cvpr} 

\usepackage{amsmath}
\usepackage{amssymb}
\usepackage{booktabs}
\usepackage{multirow}
\usepackage{indentfirst}
\usepackage[table,dvipsnames]{xcolor}

\definecolor{cvprblue}{rgb}{0.21,0.49,0.74}
\usepackage[pagebackref,breaklinks,colorlinks,citecolor=cvprblue]{hyperref}
\usepackage{float}
\def\paperID{11721} 
\def\confName{CVPR}
\def\confYear{2024}

\title{DiffAM: Diffusion-based Adversarial Makeup Transfer for Facial Privacy Protection}

\author{Yuhao Sun, Lingyun Yu\thanks{Corresponding author.},  Hongtao Xie, Jiaming Li,  Yongdong Zhang\\
University of Science and Technology of China\\
{\tt\small \{syh3327,ljmd\}@mail.ustc.edu.cn}\quad \tt\small \{yuly,htxie,zhyd73\}@ustc.edu.cn}

\begin{document}
\maketitle
\begin{abstract}
With the rapid development of face recognition (FR) systems, the privacy of face images on social media is facing severe challenges due to the abuse of unauthorized FR systems. Some studies utilize adversarial attack techniques to defend against malicious FR systems by generating adversarial examples. However, the generated adversarial examples, i.e., the protected face images, tend to suffer from subpar visual quality and low transferability. In this paper, we propose a novel face protection approach, dubbed DiffAM, which leverages the powerful generative ability of diffusion models to generate high-quality protected face images with adversarial makeup transferred from reference images. To be specific, we first introduce a makeup removal module to generate non-makeup images utilizing a fine-tuned diffusion model with guidance of textual prompts in CLIP space. As the inverse process of makeup transfer, makeup removal can make it easier to establish the deterministic relationship between makeup domain and non-makeup domain regardless of elaborate text prompts. 
Then, with this relationship, a CLIP-based makeup loss along with an ensemble attack strategy is introduced to jointly guide the direction of adversarial makeup domain, achieving the generation of protected face images with natural-looking makeup and high black-box transferability. Extensive experiments demonstrate that DiffAM achieves higher visual quality and attack success rates with a gain of \textbf{12.98\%} under black-box setting compared with the state of the arts. The code will be available at \href{https://github.com/HansSunY/DiffAM}{https://github.com/HansSunY/DiffAM}.
\end{abstract}    
\section{Introduction}
\label{sec:intro}
Recent years have witnessed major advances in face recognition (FR) systems based on deep neural networks (DNNs), which have been applied to various scenarios. However, 
the expanding capabilities of FR systems have raised concerns about the threats they pose to facial privacy. Particularly, FR systems have the potential for unauthorized surveillance and monitoring, which can analyze social media profiles without consent with the widespread availability of face images on social media\cite{besmer2010moving,smith2012big}. Therefore, it is crucial to find an effective approach to protect facial privacy against unauthorized FR systems.
\begin{figure}
  \centering
    \includegraphics[width=\linewidth]{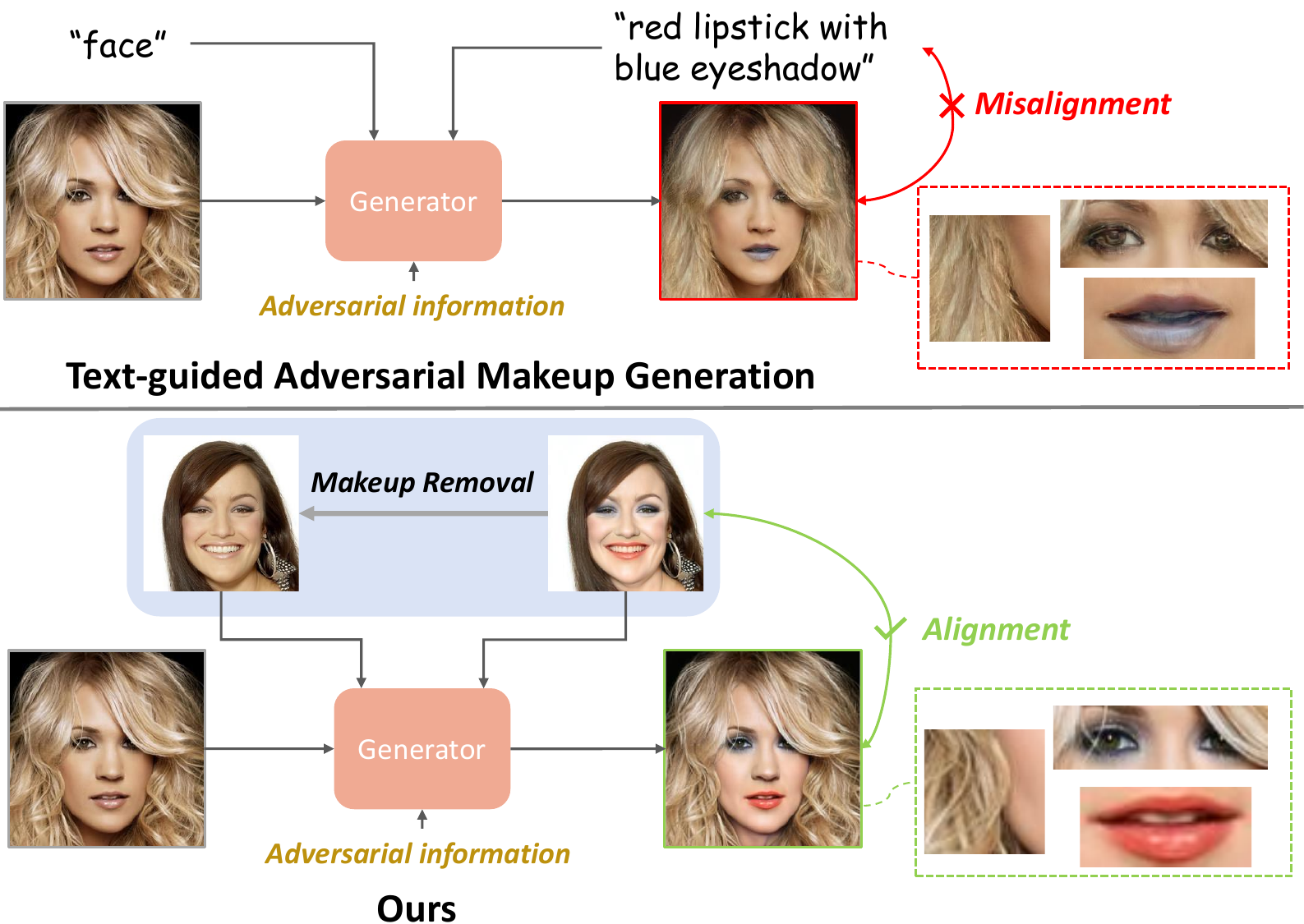}
    \caption{Core idea comparison. Text-guided method generates adversarial makeup simply with a pair of textual prompts. The coarse-grained guidance of text results in unexpected makeup generation (as shown in red boxes). Our method introduces a makeup removal module to transition this task from text-based guidance to image-based guidance and controls the direction and distance of refined adversarial makeup generation (as shown in green boxes).}
    \label{fig:comp}
\end{figure}

Many studies leverage adversarial attacks\cite{GoodfellowSS14} for facial privacy protection, which generate noise-based\cite{cherepanova2021lowkey,shan2020fawkes,yang2021towards} or patch-based\cite{komkov2021advhat,xiao2021improving} perturbations on face images. Nevertheless, to achieve ideal attack effects, most of the adversarial perturbations generated by these methods are noticeable and cluttered. Consequently, the protected face images tend to suffer from poor visual quality.
To gain more natural-looking adversarial examples, makeup-based methods\cite{Yin2021AdvMakeupAN,hu2022protecting,shamshad2023clip2protect} are attracting considerable attention. These methods organize perturbations as makeup, which can generate protected face images with adversarial makeup. However, such makeup-based methods have the following problems: (1) \textbf{Subpar visual quality.} Most of the existing works generate makeup with generative adversarial networks\cite{goodfellow2014generative} (GANs). The protected face images generated by these GAN-based approaches often have unexpected makeup artifacts and struggle to preserve attributes unrelated to makeup, such as background and hair, leading to poor visual quality. (2) \textbf{Weakness in fine-grained makeup generation.} The fine-grained information of generated makeup, like the position, color shade, range and luminosity, may not align consistently with expected makeup, especially for text-guided makeup generation method as shown in \cref{fig:comp}. (3) \textbf{Low black-box transferability.} Attack effects highly rely on robust makeup generation\cite{Lyu20233DAwareAM}. Due to the limitation of makeup generation quality proposed above, these methods suffer from low attack success rates under the black-box setting including commercial APIs. In summary, it is still challenging to simultaneously achieve satisfying makeup generation and good attack effects in black-box scenarios. 

Diffusion models\cite{ho2020denoising,song2020denoising,song2020score} have shown better performance than GANs in image generation tasks thanks to more stable training process and better coverage of image distribution\cite{dhariwal2021diffusion}. Recent works explore to guide diffusion models in CLIP\cite{radford2021learning} space with textual prompts\cite{kim2022diffusionclip,avrahami2022blended,kwon2022diffusion2,liu2023more}, demonstrating promising results. Thus, it is encouraging to utilize diffusion models to generate protected face images with both high visual quality and transferability. However, for more refined tasks like makeup transfer, textual prompts are too coarse-grained for guidance  as illustrated in \cref{fig:comp}. So it is worth considering a more fine-grained way of direction guidance in CLIP space for diffusion models.

To address the above problems, we observe that although it is hard to control the refined generation of reference makeup directly with textual prompts, the makeup of reference face image can be easily removed by a fine-tuned diffusion model with guidance of textual prompts in CLIP space\cite{kim2022diffusionclip}. Through this inverse process of makeup generation, domains of makeup and non-makeup can be definitely connected. Following this line of thought, we propose DiffAM, a novel diffusion-based adversarial makeup transfer framework to protect facial privacy. The overall pipeline of DiffAM is shown in \cref{fig:framework}. DiffAM aims to generate protected face images with adversarial makeup style transferred from a given reference image.
It is designed as two modules, a \textbf{text-guided makeup removal module} and an \textbf{image-guided makeup transfer module}. In the text-guided makeup removal module, we aim to remove the makeup of reference images, gaining the corresponding non-makeup reference images. 
This deterministic process simplifies the exploration of the makeup and non-makeup domains' relationship. Notably, the difference between the latent codes of makeup and non-makeup images of reference image in CLIP space indicates the accurate direction from non-makeup domain to makeup domain, providing alignment information for fine-grained makeup transfer. In the image-guided adversarial makeup transfer module, a CLIP-based makeup loss is proposed, combined with an ensemble attack strategy to control the precise generation direction and distance to adversarial makeup domain. In this way, high-quality makeup with strong transferability can be generated with fine-grained cross-domain guidance in CLIP space with diffusion models. 

Extensive experiments on the CelebA-HQ\cite{karras2017progressive} and LADN\cite{gu2019ladn} datasets demonstrate the effectiveness of our method in protecting facial privacy against black-box FR models
with a gain of \textbf{12.98\%}, while achieving outstanding visual quality.
In summary, our main contributions are:
\begin{itemize}
    \item A novel diffusion-based adversarial makeup transfer method, called DiffAM, is proposed for facial privacy protection, 
    intending to craft adversarial faces with high visual quality and black-box transferability.
    \item A text-guided makeup removal module is designed to establish the deterministic relationship between non-makeup and reference makeup domains, offering precise cross-domain alignment guidance for makeup transfer.
    \item A CLIP-based makeup loss is proposed for refined makeup generation. It consists of a makeup direction loss and a pixel-level makeup loss, which jointly control the direction and distance of makeup generation.
\end{itemize}
\section{Related Works}
\subsection{Adversarial Attacks on Face Recognition}
Due to the vulnerability of DNNs to adversarial examples\cite{GoodfellowSS14,szegedy2014intriguing}, many methods have been proposed to attack DNN-based face recognition (FR) systems. According to the knowledge about the target FR model, the attacks can be categorized into two main types, white-box attacks\cite{GoodfellowSS14,madry2018towards,yang2021attacks} and black-box attacks\cite{dong2018boosting,zhong2020towards,xiao2021improving,yang2021towards}.

In white-box attacks, the attacker requires complete information about the target models. However, it is hard to get full access to unauthorized FR systems in real-world scenarios. So black-box attacks, without the limitation of knowledge about the target models, are more suitable in such scenarios. Noise-based methods\cite{yang2021towards,dong2018boosting,dong2019evading}, a common form of black-box attack, can generate transferable adversarial perturbations on face images. But due to the $\ell_\infty$ constraint of noise, the attack strength cannot be guaranteed. For better attack effect, patch-based methods\cite{xiao2021improving,komkov2021advhat,sharif2019general} add abrupt adversarial patches to the limited region of face images. Although these methods attain a measure of privacy protection, the visual quality of the resulting protected face images is often compromised and suffers from weak transferability. Recent works attempt to protect face images with adversarial makeup\cite{hu2022protecting,shamshad2023clip2protect,Yin2021AdvMakeupAN}, which is an ideal solution for balancing visual quality and transferability. These methods hide the adversarial information in the generated makeup style, which can fool FR systems in an imperceptive way. However, existing makeup-based methods tend to suffer from poor visual quality and low transferability. And the attributes unrelated to makeup are hard to be completely preserved. Therefore, in this work, we propose a novel face protection approach DiffAM to improve the quality and black-box transferability of adversarial makeup through fine-grained guidance in CLIP space.
\subsection{Makeup Transfer}
Makeup transfer\cite{li2018beautygan,gu2019ladn,jiang2020psgan,chen2019beautyglow,chang2018pairedcyclegan,deng2021spatially} aims to transfer makeup styles from the reference faces to the source faces while preserving the original face identity. As a typical image-to-image translation task, many approaches employ generative adversarial networks (GANs) for makeup transfer. BeautyGAN\cite{li2018beautygan} first introduces a dual input/output GAN to achieve makeup transfer and removal simultaneously. Moreover, it proposes a pixel-wise histogram matching loss as guidance for makeup transfer in different face regions which has been subsequently adopted by many methods. LADN\cite{gu2019ladn} adopts multiple overlapping local discriminators and asymmetric losses for heavy facial makeup transfer. 
Besides the above GAN-based methods, BeautyGlow\cite{chen2019beautyglow} uses the Glow framework for makeup transfer by decomposing the latent vector into non-makeup and makeup parts. Taking advantage of the good visual properties of makeup transfer, we apply the concept of makeup style to facial privacy protection. The proposed DiffAM organizes the distribution of adversarial information semantically into adversarial makeup, which can minimize the impact on the visual quality of the protected face images while ensuring the effectiveness of attacks on the FR system. 
\subsection{Diffusion model and Style Transfer}   
Diffusion models\cite{nichol2021improved,dhariwal2021diffusion,ho2020denoising,sohl2015deep} are a class of probabilistic generative models, which have impressive performance in generating high-quality images. They have been applied to various tasks, such as image generation\cite{dhariwal2021diffusion,song2020denoising}, image editing\cite{couairon2023diffedit,kim2022diffusionclip,avrahami2022blended}, image super-resolution\cite{li2022srdiff,gao2023implicit} and style transfer\cite{kim2022diffusionclip,zhang2023inversion,kwon2022diffusion2}. Style transfer is an image-to-image translation\cite{isola2017image,zhu2017unpaired} task that combines the content of source image and the style of reference image. Existing diffusion-based methods leverage the alignment between text and images in CLIP space for text-driven style transfer\cite{kim2022diffusionclip,kwon2022diffusion2}. As a subtask of style transfer, makeup transfer can also be guided with text. However, the control of text is too rough for more refined makeup transfer in comparison to global style transfer. Given a reference makeup image, simply using text is insufficient to generate precise makeup, such as the intensity, shape, and position. Considering the limitation of text, we point that makeup removal, the inverse process of makeup transfer, can provide deterministic guidance from non-makeup domain to makeup domain. Moreover, a CLIP-based makeup loss is introduced for image-driven makeup transfer.  
\begin{figure*}
\setlength{\belowcaptionskip}{-0.5cm}
  \centering
    \includegraphics[width=\linewidth]{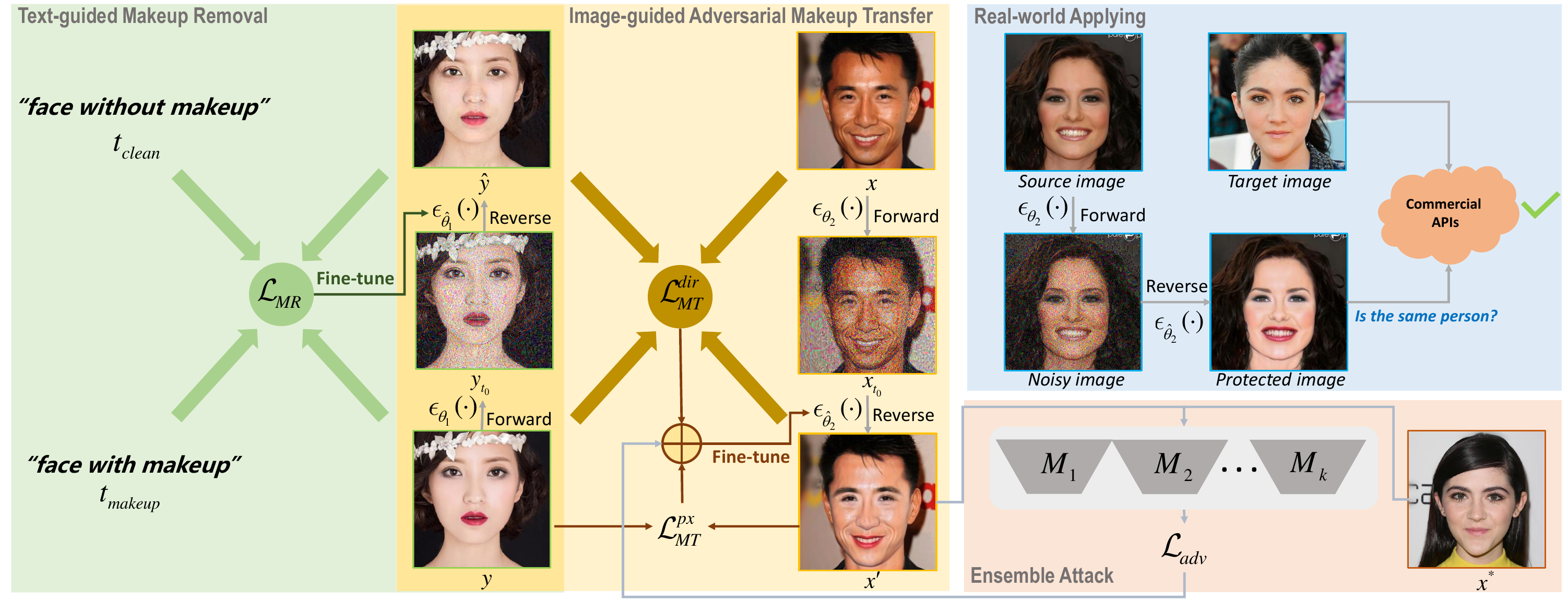}
    \caption{Overview of DiffAM. DiffAM is a two-stage framework that generates protected face image $x^\prime$ by transferring the makeup style of $y$ to $x$. Specifically, in text-guided makeup removal module, we input a reference image $y$ and obtain the non-makeup $\hat{y}$ through text guidance, determining the precise makeup direction. Then, in image-guided adversarial makeup transfer module, we input a face image $x$ and obtain the adversarial-makeup image $x^\prime$ through image guidance of $y$ and $\hat{y}$, along with an ensemble attack strategy. }
    \label{fig:framework}

\end{figure*}
\section{Method}
\subsection{Problem Formulation}
Black-box attacks on face recognition (FR) systems can be further divided into targeted attacks ($i.e.$, impersonation attacks) and non-targeted attacks ($i.e.$, dodging attacks). For more efficient protection of face images, \textit{we focus on targeted attack which aims to mislead FR systems to recognize the protected faces as the specified target identity.} The targeted attack can be defined as an optimization problem:
\begin{equation}
  \min_{x^{\prime}} L_{adv}=\mathcal{D}(m(x^{\prime}),m(x^*)),
  \label{eq:9}
\end{equation}
where $x^{\prime}$ is the protected face image, $x^*$ is the target face image, $m$ represents the feature extractor of FR models, $\mathcal{D}(.)$ represents a distance metric. 

Particularly, as for adversarial makeup transfer, the protected face image $x^{\prime}$ is obtained by transferring makeup style from the reference image $y$ to the clean face image $x$, which can be formulated as:
\begin{equation}
  x^{\prime}=\mathcal G(x,y),
  \label{eq:10}
\end{equation}
where $\mathcal{G}$ is an adversarial makeup transfer network. 
\subsection{DiffAM}
To generate natural-looking and transferable adversarial makeup against FR models, DiffAM aims to explore precise and fine-grained guidance of generation from non-makeup domain to adversarial makeup domain, which is the overlap domain between reference makeup domain and adversarial domain, as shown in \cref{fig:dom}(a). To achieve this, DiffAM consists of two stages: text-guided makeup removal and image-guided adversarial makeup transfer, as illustrated in \cref{fig:framework}. The details of each component are described as follows.
\subsubsection{Text-guided Makeup Removal}
\begin{figure*}
\setlength{\belowcaptionskip}{-0.5cm}
  \centering

    \includegraphics[width=\linewidth]{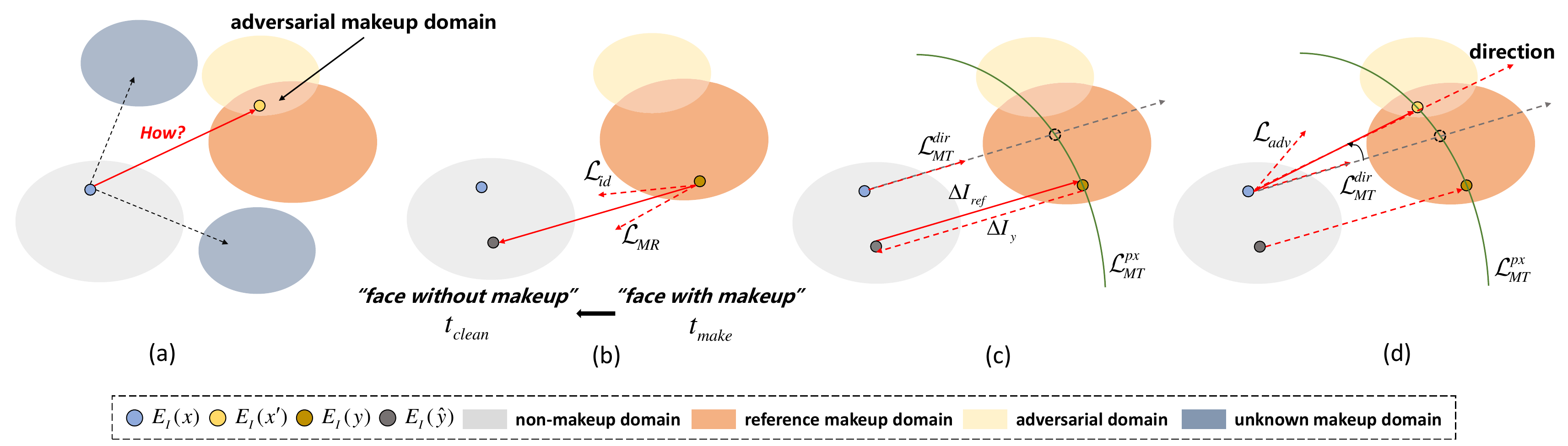}
    \caption{The process of adversarial makeup transfer in CLIP space. (a) It is challenging to directly find a precise path from the non-makeup domain to the adversarial makeup domain. (b) The process of text-guided makeup removal can help establish the relationship between domains. (c) The inverse direction of makeup removal indicates the direction to makeup domain for makeup transfer and the pixel-level makeup loss guides the distance to makeup domain. (d) The direction of ensemble attack and makeup transfer jointly guide the final direction to the adversarial makeup domain.}
    \label{fig:dom}

\end{figure*}

\noindent Adopting makeup transfer against FR models, an intuitive idea is to directly use text pairs to guide the generation of makeup. However, the coarse-grained text is hard to build the precise relationship between non-makeup domain and reference makeup domain. Concretely, the details of reference makeup, such as color depth, shape, etc., are difficult to control in text, resulting in undesired makeup generation. To eliminate ambiguity caused by textual guidance, we innovatively design the text-guided makeup removal module 
to remove the makeup style of reference image $y$ and obtain the corresponding non-makeup image $\hat y$ with the guidance of text pair in CLIP space. The pair of reference images with and without makeup can connect makeup domain and non-makeup domain in CLIP space, as illustrated in \cref{fig:dom}(b), providing the deterministic cross-domain guidance for subsequent stage of adversarial makeup transfer.

Given the reference image $y$, we first convert it to the latent $y_{t_0}$ by forward diffusion process with a pre-trained diffusion model $\epsilon_{\theta_1}$. In the reverse diffusion process, the diffusion model $\epsilon_{\theta_1}$ is fine-tuned for makeup removal to obtain the non-makeup image $\hat{y}$, which is guided by directional CLIP loss $\mathcal L_{\textit{MR}}$ \cite{kim2022diffusionclip}:
\begin{equation}
  \mathcal L_{\textit{MR}}=1-\frac{\Delta I_y\cdot\Delta T}{\Vert\Delta I_y\Vert\Vert\Delta T\Vert},
  \label{eq:11}
\end{equation}
where $\Delta I_y=E_I(\hat{y}(\hat{\theta}_1))-E_I(y)$ and $\Delta T=E_T(t_{clean})-E_T(t_{makeup})$. Here, $E_I$ and $E_T$ are the image and text encoders of CLIP model\cite{radford2021learning}, $\hat{y}(\hat{\theta}_1)$ is the sampled image from $y_{t_0}$ with the optimized parameter $\hat{\theta}_1$, $t_{clean}$ and $t_{makeup}$ are the text descriptions for non-makeup and makeup domains, which can be simply set as \textit{"face without makeup"} and \textit{"face with makeup"}.

To preserve identity information and image quality, we introduce the face identity loss $\mathcal L_{id}(\hat{y},y)$ \cite{deng2019arcface} and perceptual loss $\mathcal L_{\textit{LPIPS}}(\hat{y},y)$ \cite{zhang2018unreasonable}. As for total loss $\mathcal L_{total}$, we have
\begin{equation}
  \mathcal L_{total}=\lambda_{\textit{MR}}\mathcal L_{\textit{MR}}+\lambda_{id}\mathcal L_{id}+\lambda_{\textit{LPIPS}}\mathcal L_{\textit{LPIPS}},
  \label{eq:12}
\end{equation}
where $\lambda_{\textit{MR}}$, $\lambda_{id}$ and $\lambda_{\textit{LPIPS}}$ are weight parameters.

It is worth noting that we use deterministic DDIM sampling and DDIM inversion\cite{song2020denoising} as the reverse diffusion process and forward diffusion process. The reconstruction capability of deterministic DDIM inversion and sampling ensures the effects of makeup removal.
\subsubsection{Image-guided Adversarial Makeup Transfer}
\label{sec:transfer}
\noindent To protect source image $x$ against FR models, the image-guided adversarial makeup transfer module aims to generate protected face image $x^{\prime}$ with adversarial makeup transferred from reference image $y$. The protected face image $x^{\prime}$ misleads FR models into recognizing it as the target identity $x^*$, as shown in \cref{fig:framework}. After getting $y$ and $\hat{y}$, a CLIP-based makeup loss $\mathcal L_{\textit{MT}}$ coordinating with an ensemble attack strategy is introduced to align the direction between $x$ and $x^{\prime}$ with the direction between non-makeup domain and adversarial makeup domain. The fine-grained cross-domain alignment ensures the quality and black-box transferability of adversarial makeup style.

Given the source image $x$, we first get the latent $x_{t_0}$ through deterministic DDIM inversion with another pre-trained diffusion model $\epsilon_{\theta_2}$. Then, the diffusion model $\epsilon_{\theta_2}$ is fine-tuned to generate protected face image $x^{\prime}$ with guidance of CLIP-based makeup loss $\mathcal L_{\textit{MT}}$ and ensemble attack loss $L_{adv}$. We also incorporate a makeup-irrelevant information preservation operation for better visual quality during fine-tuning. The details of the fine-tuning process are presented as follows.

\noindent \textbf{CLIP-based Makeup Loss.} During the stage of makeup removal, the learned direction from reference makeup domain to non-makeup domain in CLIP space is expressed as $\Delta I_y = E_I(\hat{y})-E_I(y)$. As shown in \cref{fig:dom}(c), We can just reverse the direction to get the guidance of direction from non-makeup domain to reference makeup domain:
\begin{equation}
    \Delta I_{ref} = -\Delta I_y = E_I(y)-E_I(\hat{y}),
  \label{eq:13}
\end{equation}
where $E_I$ is the image encoder of CLIP model. In addition to maintaining consistency with the style removal stage in CLIP space, $E_I$ has powerful image understanding capabilities, which can facilitate better extraction of semantic information of makeup styles, such as shape and the relative position on the face\cite{gal2022stylegan,kim2022diffusionclip}. In this way, $\Delta I_{ref}$ can achieve more semantic and holistic supervision than simple pixel-level guidance or text guidance. To align $\Delta I_{x}$, the direction between $x$ and $x^{\prime}$, with $\Delta I_{ref}$ in CLIP space, a makeup direction loss is proposed:
\begin{equation}
     \mathcal L_{\textit{MT}}^{dir}=1-\frac{\Delta I_x\cdot\Delta I_{ref}}{\Vert\Delta I_x\Vert \Vert\Delta I_{ref}\Vert},
  \label{eq:14}
\end{equation}
where $\Delta I_x = E_I(x^\prime(\hat{\theta}_2)) - E_I(x)$ and $x^\prime(\hat{\theta}_2)$ is the protected face image generated by fine-tuned diffusion model $\epsilon_{\hat{\theta}_2}$. By aligning image pairs in CLIP space, makeup direction loss controls precise direction for makeup transfer.

Besides the guidance of makeup transfer direction, we also need to consider the makeup transfer distance between makeup domain and non-makeup domain, as shown in \cref{fig:dom}(c), which determines the intensity and accurate color of makeup. Therefore, a pixel-level makeup loss $L_{\textit{MT}}^{px}$\cite{li2018beautygan} is employed to constrain makeup transfer distance in pixel space. We conduct histogram matching between generated image $x^\prime$ and reference image $y$ on three facial regions as guidance for the intensity of makeup. The pixel-level makeup loss is defined as:
\begin{equation}
     \mathcal L_{\textit{MT}}^{px} = \Vert x^\prime-HM(x^\prime,y)\Vert,
  \label{eq:15}
\end{equation}
where $HM(.)$ represents the histogram matching.

Combining the makeup direction loss and pixel-level makeup loss, the CLIP-based makeup loss is expressed as: 
\begin{equation}
     \mathcal L_{\textit{MT}} = \lambda_{dir} \mathcal L_{\textit{MT}}^{dir}+\lambda_{px}\mathcal L_{\textit{MT}}^{px}.
  \label{eq:16}
\end{equation}
With the joint guidance of $\mathcal L_{\textit{MT}}^{dir}$ and $\mathcal L_{\textit{MT}}^{px}$ for makeup transfer direction and distance, the generated makeup image $x^\prime$ can precisely fall within the reference makeup domain, achieving excellent makeup transfer effects.

\noindent \textbf{Ensemble Attack.}
In addition to guidance in makeup transfer direction, there is also a need for guidance in the adversarial direction to find the final adversarial makeup domain, as shown in \cref{fig:dom}(d). To solve the optimization problem in \cref{eq:9}, an ensemble attack strategy\cite{hu2022protecting} is introduced. We choose $K$ pre-trained FR models with high recognition accuracy as surrogate models for fine-tuning, aiming to find the direction towards a universal adversarial makeup domain. The ensemble attack loss is formulated as:
\begin{equation}
     \mathcal L_{adv} = \frac{1}{K}\sum^{K}_{k=1}[1-\cos (m_k(x^\prime),m_k(x^*))],
  \label{eq:17}
\end{equation}
where $m_k$ represents the $k$-th pre-trained FR model and we use cosine similarity as the distance metric.

The ensemble attack loss $\mathcal L_{adv}$ adjusts the generation direction from the makeup domain to the adversarial makeup domain, improving the transferability of adversarial makeup under black-box settings. 

\noindent \textbf{Preservation of Makeup-Irrelevant Information.} To ensure the visual quality of protected face images, it is crucial to minimize the impact on makeup-irrelevant information, such as identity and background, during makeup transfer. However, due to fine-tuning the diffusion model $\epsilon_{\theta_2}$ in the sampling process, the cumulative error between the prediction noise of $\epsilon_{\theta_2}$ and $\epsilon_{\hat{\theta}_2}$ will increase with denoising steps, resulting some unexpected distortion besides makeup style. To address this problem, we leverage the progressive generation property of diffusion models\cite{ho2020denoising,meng2021sdedit}, where coarse-grained information ($e.g.$, layout, shape) is focused at early denoising steps while semantic details at later steps. Makeup style is typically generated in the final steps of denoising process as a kind of fine-grained information of face. Thus, we propose to reduce the time step $T$ in DDIM inversion and sampling for retention of most makeup-irrelevant information. This simple but effective operation can greatly improve the visual quality of protected face and accelerate the whole process of makeup transfer.

Moreover, the perceptual loss $\mathcal L_{\textit{LPIPS}}(x^\prime,x)$ and $\ell_1$ loss are further introduced to explicitly control generation quality and pixel similarity:
\begin{equation}
     \mathcal L_{vis} = \mathcal L_{\textit{LPIPS}}(x^\prime,x) + \lambda_{\ell_1}\Vert x^\prime-x\Vert.
  \label{eq:18}
\end{equation}

\noindent \textbf{Total Loss Function}. By combining all the above loss functions, we have total loss function as follows:
\begin{equation}
     \mathcal L = \lambda_{\textit{MT}}\mathcal L_{\textit{MT}}+\lambda_{adv} \mathcal L_{adv}+\lambda_{vis}\mathcal L_{vis},
  \label{eq:19}
\end{equation}
where $\lambda_{\textit{MT}}$, $\lambda_{adv}$ and $\lambda_{vis}$ are weight parameters. 
\begin{figure*}
\setlength{\belowcaptionskip}{-0.3cm}
  \centering
    \includegraphics[width=0.865\linewidth]{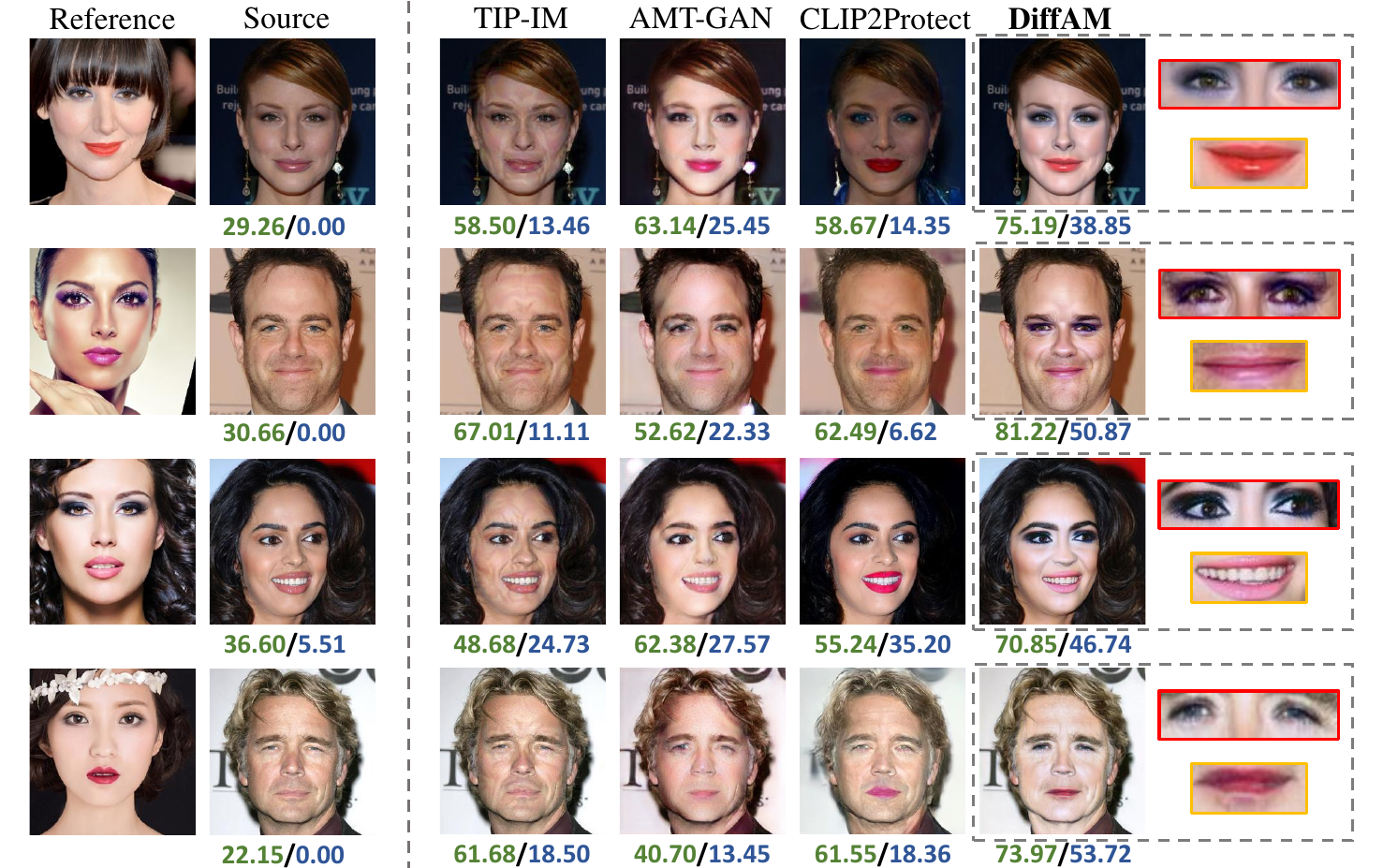}
    \caption{Visualizations of the protected face images generated by different facial privacy protection methods on CelebA-HQ. The green and blue numbers below each image are confidence scores returned by Face++ and Aliyun.}
    \label{fig:vis}
\end{figure*}
\begin{table*}[htbp]
\setlength{\belowcaptionskip}{-0.5cm}
  \centering
  \resizebox{\linewidth}{!}{
  \begin{tabular}{c|c|cccc|cccc|c}
  \hline
  &\multirow{2}{*}{Method}&\multicolumn{4}{c|}{CelebA-HQ}&\multicolumn{4}{c|}{LADN-dataset}&\multirow{2}{*}{Average}\\
  \cline{3-10}
  &&IRSE50&IR152&Facenet&Mobileface&IRSE50&IR152&Facenet&Mobileface\\
    \cline{2-11}
    &Clean & 7.29 & 3.80 & 1.08 & 12.68&2.71&3.61&0.60&5.11&4.61\\
    \hline
    \multirow{4}{*}{Noise-based}&PGD & 36.87 &20.68&1.85&43.99&40.09&19.59&3.82&41.09&25.60 \\
    &MI-FGSM & 45.79&25.03&2.58&45.85&48.90&25.57&6.31&45.01&30.63\\
    &TI-DIM & 63.63&36.17&15.30&\cellcolor{orange!20}57.12&56.36&34.18&22.11&48.30&41.64\\
    &TIP-IM&54.40&\cellcolor{orange!20}37.23&\cellcolor{orange!20}40.74&48.72&65.89&43.57&\cellcolor{orange!50}63.50&46.48&50.06\\
    \cline{1-11}
    \multirow{4}{*}{Makeup-based}&Adv-Makeup&21.95&9.48&1.37&22.00&29.64&10.03&0.97&22.38&14.72\\
    &AMT-GAN&\cellcolor{orange!20}76.96&35.13&16.62&50.71&\cellcolor{orange!20}89.64&\cellcolor{orange!20}49.12&32.13&\cellcolor{orange!20}72.43&\cellcolor{orange!20}52.84\\
&CLIP2Protect&\cellcolor{orange!50}81.10&\cellcolor{orange!50}48.42&\cellcolor{orange!50}41.72&\cellcolor{orange!50}75.26&\cellcolor{orange!50}91.57&\cellcolor{orange!50}53.31&\cellcolor{orange!20}47.91&\cellcolor{orange!50}79.94&\cellcolor{orange!50}64.90\\
    \cline{2-11}
&Ours&\cellcolor{orange!90}92.00&\cellcolor{orange!90}63.13&\cellcolor{orange!90}64.67&\cellcolor{orange!90}83.35&\cellcolor{orange!90}95.66&\cellcolor{orange!90}66.75&\cellcolor{orange!90}65.44&\cellcolor{orange!90}92.04&\cellcolor{orange!90}77.88\\
    \hline
  \end{tabular}
  }
  \caption{Evaluations of \textit{attack success rate} (ASR) for black-box attacks. For each column, we choose the other three FR models as surrogates to generate protected face images. DiffAM achieves a \textbf{12.98\%} improvement on average ASR compared to the state of the art.}
  \label{tab:asr}
\end{table*}
\begin{table}[htbp]
\setlength{\belowcaptionskip}{-0.7cm}
    \centering
    \begin{tabular}{c|ccc}
    \hline
    &FID($\downarrow$)&PSNR($\uparrow$)&SSIM($\uparrow$)\\
    \hline
    Adv-makeup&4.2282&34.5152&0.9850  \\
    AMT-GAN&34.4405&19.5045&0.7873\\
    CLIP2Protect&37.1172&19.3537&0.6025\\
    \hline
    DiffAM (w/o $L_{\textit{MT}}^{dir}$)&33.6896&19.3099&0.8651\\
    DiffAM ($T=200$)&47.3186&18.6768&0.8367\\
    DiffAM ($T=100$)&32.4767&19.8816&0.8742\\
    Our DiffAM 
    &26.1015&20.5260&0.8861\\
    \hline
    \end{tabular}
    \caption{Quantitative evaluations of image quality. Our DiffAM  represents the results with $L_{\textit{MT}}^{dir}$ and $T=60$.}
    \label{tab:vis_qiality}
\end{table}
\section{Experiments}
\subsection{Experimental Settings}
\noindent \textbf{Datasets.} For makeup removal, 
we randomly sample 200 makeup images from MT dataset\cite{li2018beautygan}, which consists of 2719 makeup images and 1115 non-makeup images, for fine-tuning. For adversarial makeup transfer, we randomly sample 200 images from CelebA-HQ dataset\cite{karras2017progressive} for fine-tuning. To evaluate the effectiveness of DiffAM, we choose CelebA-HQ and LADN\cite{gu2019ladn} as our test sets. For CelebA-HQ, we select a subset of 1000 images and divide them into four groups, each of which has a target identity\cite{hu2022protecting}. Similarly, for LADN, we divide the 332 images into four groups for attack on different target identities. 

\noindent \textbf{Benchmark.} We do comparisons with multiple benchmark schemes of adversarial attacks, including PGD\cite{madry2018towards}, MI-FGSM\cite{dong2018boosting}, TI-DIM\cite{dong2019evading}, TIP-IM\cite{yang2021towards}, Adv-Makeup\cite{Yin2021AdvMakeupAN}, AMT-GAN\cite{hu2022protecting} and CLIP2Protect\cite{shamshad2023clip2protect}. PGD, MI-FGSM, TI-DIM and TIP-IM are typical noise-based methods, while Adv-Makeup, AMT-GAN, CLIP2Protect and DiffAM are makeup-based methods that also exploit makeup transfer to generate protected face images. 

\noindent \textbf{Target Models.} We choose four popular public FR models as the attacked models, including IR152\cite{deng2019arcface}, IRSE50\cite{hu2018squeeze}, FaceNet\cite{schroff2015facenet} and MobileFace\cite{chen2018mobilefacenets}. Three of them are chosen for ensemble attack during training and the remaining one serves as the black-box model for testing. Meanwhile, we evaluate the performance of DiffAM on commercial FR APIs including Face++\footnote{\url{https://www.faceplusplus.com/face-comparing/}} and Aliyun\footnote{\url{https://vision.aliyun.com/experience/detail?&tagName=facebody&children=CompareFace}}. 

\noindent \textbf{Implemention Details.} For text-guided makeup removal and image-guided makeup transfer, we use ADM\cite{dhariwal2021diffusion} pre-trained on Makeup Transfer (MT) dataset\cite{li2018beautygan} and CelebA-HQ dataset\cite{karras2017progressive} respectively as the generative model. To fine-tune diffusion models, we use an Adam optimizer\cite{KingBa15} with an initial learning rate of 4e-6. It is increased linearly by 1.2 per 50 iterations. As mentioned in \cref{sec:transfer}, we set total time step $T = 60$ and ($S_{inv}$, $S_{sam}$) = (20, 6), where $S_{inv}$ and $S_{sam}$ represent the discretization steps of DDIM inversion and sampling. The diffusion models are fine-tuned with 6 epochs. All our experiments are conducted on one NVIDIA RTX3090 GPU.

\noindent \textbf{Evaluation Metrics.} Following \cite{hu2022protecting,shamshad2023clip2protect}, we use attack success rate (ASR) to evaluate the effectiveness of privacy protection of different methods. When calculating the ASR, we set False Acceptance Rate (FAR) at 0.01 for each FR model. In addition, we use FID\cite{heusel2017gans}, PSNR(dB) and SSIM\cite{wang2004image} to evaluate the image quality of protected face images.
\subsection{Comparison Study}
This section compares the experimental results of DiffAM and benchmark methods in terms of attack performance under black-box settings and image quality.
\begin{figure*}
\setlength{\belowcaptionskip}{-0.55cm}
  \centering
    \includegraphics[width=\linewidth]{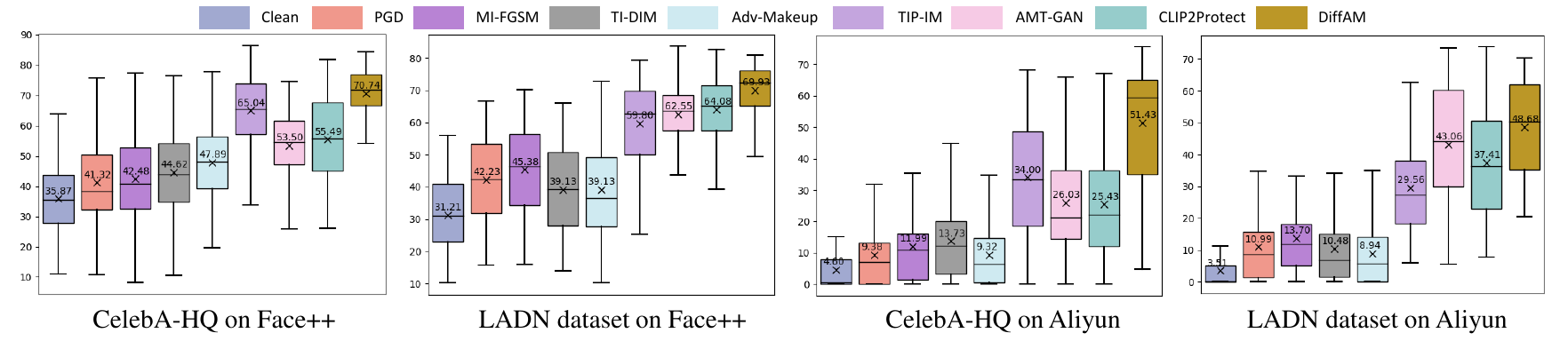}
    \caption{The confidence scores (higher is better) returned from commercial APIs, Face++ and Aliyun. DiffAM has higher and more stable confidence scores than state-of-the-art noise-based and makeup-based facial privacy protection methods.}
    \label{fig:api_comp}
\end{figure*}
\begin{figure}
\setlength{\belowcaptionskip}{-0.5cm}
  \centering
    \includegraphics[width=\linewidth]{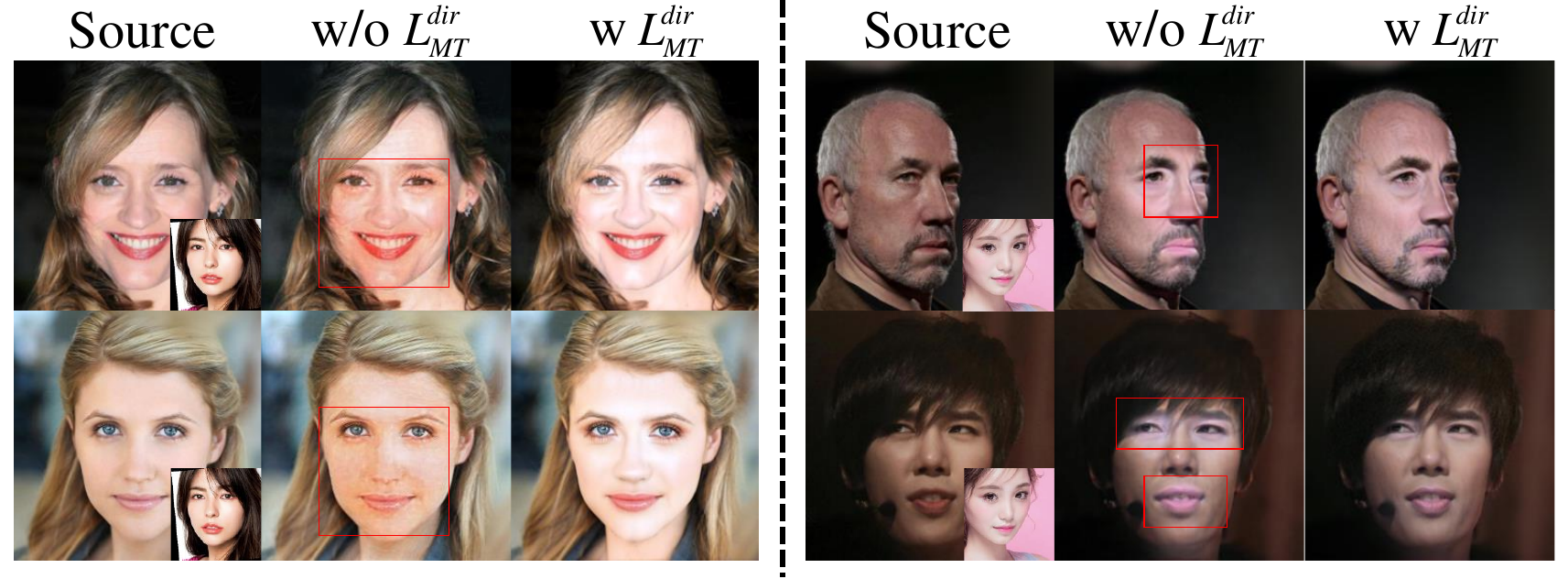}
    \caption{Ablation study for the makeup direction loss. The generated protected face images without makeup direction loss have obvious makeup artifacts (red boxes).}
    \label{fig:direction}
\end{figure}

\noindent \textbf{Comparison on black-box attacks.} \cref{tab:asr} reports quantitative results of black-box attacks against four popular FR models on CelebA-HQ and LADN datasets. We test the performance of targeted attack against four target identities\cite{hu2022protecting}, with the results of DiffAM averaged over 5 reference makeup images from MT-dataset, following \cite{shamshad2023clip2protect}. To simulate real-world protection scenarios, the target face images used during testing are different images of the same individual compared to the one used during training. The average black-box ASRs of DiffAM are significantly about 28\% and 13\% higher than SOTA noise-based method TIP-IM and makeup-based method CLIP2Protect. DiffAM also maintains a good attack effectiveness on Facenet, which is difficult to attack using other methods. The results show that DiffAM has strong black-box transferability, which demonstrates the role of DiffAM in accurate guidance to the adversarial makeup domain as we expected. 

\noindent \textbf{Comparison on image quality.} \cref{tab:vis_qiality} reports the evaluations of image quality. We choose Adv-makeup, AMT-GAN and CLIP2Protect, three latest makeup-based methods, as benchmarks for comparison. Adv-makeup has the best performance in all quantitative assessments. This is because Adv-makeup only generates eyeshadow compared to the full-face makeup generation of the others. Although Adv-makeup has minimal image modification, the trade-off is a significantly lower attack success rate as shown in \cref{tab:asr}. Compared to AMT-GAN and CLIP2Protect, DiffAM achieves lower FID scores and higher PSNR and SSIM scores, which indicates that the adversarial makeups generated by DiffAM are more natural-looking and have less impact on images at the pixel level. 

We also show the qualitative comparison of visual quality in \cref{fig:vis}. Note that for text-guided method CLIP2Protect, we use textual prompts, such as \textit{“purple lipstick with purple eyeshadow”}, derived from the reference images to generate makeup. Compared to the noise-based method TIP-IM, DiffAM generates more natural-looking protected face images without noticeable noise patterns. As for makeup-based methods, AMT-GAN fails to transfer makeup precisely and the generated face images have obvious makeup artifacts. CLIP2Protect struggles to generate accurate makeup corresponding to the given textual prompt and loses most of image details. In contrast, DiffAM stands out for accurate and high-quality makeup transfer, such as lipstick and eyeshadow, 
thanks to fine-grained supervision of generation direction and distance. Our proposed operation for preserving makeup-irrelevant information also ensures that face image details are well-preserved. Notably, DiffAM is effective in generating makeup for male images, which is a challenge for other makeup-based methods.
\subsection{Attack Performance on Commercial APIs}
\cref{fig:api_comp} shows the quantitative results of attacks on commercial APIs Face++ and Aliyun. We randomly selected 100 images each from CelebA-HQ and LADN datasets to protect and report confidence scores returned from APIs. The confidence scores are between 0 to 100, where the higher score indicates higher similarity between the protected face image and the target image. DiffAM achieves the highest average confidence scores about 70 and 50 on each API and the attack effect is relatively stable across different datasets, which indicates the strong black-box attack capability in real-world scenarios.
\subsection{Ablation Studies}
\noindent \textbf{Control of Makeup Direction.}
We verify the importance of makeup direction loss $L_{\textit{MT}}^{dir}$ for makeup quality in \cref{fig:direction}. In the absence of $L_{\textit{MT}}^{dir}$, the generated makeup has obvious makeup artifacts (red boxes in \cref{fig:direction}), leading to a decrease in image quality. \cref{tab:vis_qiality} also illustrates that the generated images with $L_{\textit{MT}}^{dir}$ have better quantitative results than the ones without $L_{\textit{MT}}^{dir}$. This is because, without $L_{\textit{MT}}^{dir}$, the generated makeup is only guided by pixel-level makeup loss $L_{\textit{MT}}^{px}$. $L_{\textit{MT}}^{px}$ just supervises makeup generation in different facial segmentation regions individually without global semantic supervision, resulting in inaccurate makeup generation. By applying makeup direction loss $L_{\textit{MT}}^{dir}$, precise guidance can be provided for the global generation direction of the makeup, ensuring high-quality and accurate makeup.

\begin{figure}
\setlength{\abovecaptionskip}{-0.001cm}
\setlength{\belowcaptionskip}{-0.2cm}
  \centering
    \includegraphics[width=\linewidth]{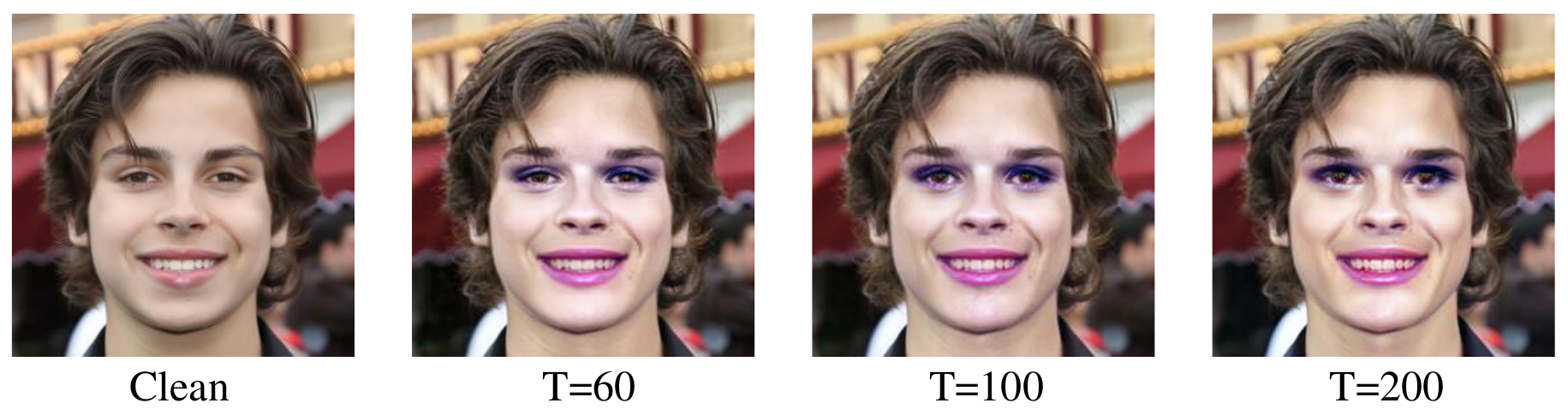}
    \caption{Impact of inversion steps on preservation of makeup-irrelevant information. As the number of inversion step $T$ increases, the facial features of the protected face image will change.}
    \label{fig:step}
\end{figure}

\noindent \textbf{Preservation of Makeup-Irrelevant Information.}
\cref{fig:step} shows the generated face images under a set of increasing inversion steps. With the increase of DDIM inversion steps, the generated face image has unexpected changes in facial attribute information. \cref{tab:vis_qiality} shows the quantitative results at different steps, indicating that it is a simple but effective operation to preserve makeup-irrelevant information by controlling DDIM inversion steps. 
\begin{table}[tb]
\setlength{\belowcaptionskip}{-0.45cm}
    \centering
    \resizebox{\linewidth}{!}{
    \begin{tabular}{c|cccccc}
    \hline
    Reference&XMY-060&vHX570&vFG137&vRX189&XYH-045&Std.\\
    \hline
    ASR&76.01&78.56&77.91&79.06&78.23&1.04  \\
    \hline
    \end{tabular}
    }
    \caption{Impact of different reference makeup styles on ASR. Five reference makeup styles are selected from MT-dataset. Std. denotes standard deviation.}
    \label{tab:diff_makeup}
\end{table}

\noindent \textbf{Robustness on different makeup styles.}
Being able to generate protected face images with any given reference makeup holds more practical value. Thus, we randomly select five reference images from MT-dataset to evaluate the impact of different reference makeup styles on attack effects of DiffAM. As shown in \cref{tab:diff_makeup}, the change of makeup styles has limited influence on ASR, which indicates the robustness of DiffAM to the change of makeup styles.  
\section{Conclusion}
In this paper, we introduce DiffAM, a novel diffusion-based adversarial makeup transfer method for facial privacy protection. Building upon the generative capabilities of diffusion models, We innovatively introduce a makeup removal module to address uncertainty in text-guided generation. The deterministic cross-domain relationship can be obtained during makeup removal process, enabling fine-grained alignment guidance for adversarial makeup generation with the proposed CLIP-based makeup loss and ensemble attack strategy. Numerous experiments have verified that DiffAM ensures strong black-box attack capabilities against many FR models and commercial APIs, while maintaining high-quality and precise makeup generation.
{
    \small
    \bibliographystyle{ieeenat_fullname}
    \bibliography{main}
}
\clearpage
\setcounter{page}{1}
\maketitlesupplementary
\section{Background: DDPM and DDIM}
Denoising Diffusion Probabilistic Model (DDPM)\cite{ho2020denoising,sohl2015deep} consists of a forward diffusion process and a reverse diffusion process. The forward diffusion process is described as a Markov chain where Gaussian noise is gradually added to the original image $x_0$ to get the noisy image $x_t$ at every time steps $t\in\{1,...,T\}$:
\begin{equation}
  q(x_t|x_{t-1})=\mathcal N(\sqrt{1-\beta_t}x_{t-1},\beta_t\mathbf{I}),
  \label{eq:1}
\end{equation}
where $\beta_t \in (0,1) $ are hyperparameters representing the variance schedule. A good property of this formulation is that we can directly sample $x_t$ given $x_0$:
\begin{equation}
  q(x_t|x_0)=\mathcal N(\sqrt{\alpha_t}x_0,(1-\alpha_t)\mathbf{I}),
  \label{eq:2}
\end{equation}
\begin{equation}
  x_t=\sqrt{\alpha_t}x_0+\sqrt{1-\alpha_t}\epsilon,\quad\epsilon\sim\mathcal N(0,\mathbf{I}),
  \label{eq:3}
\end{equation}
where $\alpha_t=\prod_{s=1}^t(1-\beta_s)$. 

We can get a new sample from the distribution $q(x_0)$ by following the reverse steps $q(x_{t-1}|x_t)$, starting from $x_T\sim\mathcal N(0,\mathbf{I})$. As the posteriors $q(x_{t-1}|x_t)$ is intractable, in the reverse process, a neural network $p_\theta$ is trained to approximate it:
\begin{equation}
  p_\theta(x_{t-1}|x_t)=\mathcal{N}(\mu_\theta(x_t,t),\sigma_t^2\mathbf{I}),
  \label{eq:4}
\end{equation}
where
\begin{equation}
  \mu_\theta(x_t,t)=\frac{1}{\sqrt{1-\beta_t}}\left(x_t-\frac{\beta_t}{1-\alpha_t}\epsilon_\theta(x_t,t)\right).
  \label{eq:5}
\end{equation}
A U-net\cite{ronneberger2015u} is trained to learn a function $\epsilon_\theta(x_t,t)$ to predict the added noise at time step $t$ by optimizing the objective\cite{ho2020denoising}:
\begin{equation}
  \min\limits_{\theta}\mathcal{L}(\theta)=\mathbb{E}_{x_0\sim q(x_0),\epsilon\sim \mathcal N(0,\mathbf{I}),t}||\epsilon-\epsilon_\theta(x_t,t)||_2^2.
  \label{eq:6}
\end{equation}
Then, we can sample the data as follows:
\begin{equation}
  x_{t-1}=\mu_\theta(x_t,t)+\sigma_tz,
  \label{eq:7}
\end{equation}
where $z\sim\mathcal N(0,\mathbf{I})$.

To accelerate the sampling process, Song \etal\cite{song2020denoising} proposed Denoising Diffusion Implicit Model (DDIM) that has a non-Markovian noising process. The sampling process of DDIM is:
\begin{equation}
  x_{t-1}=\sqrt{\alpha_{t-1}}f_\theta(x_t,t)+\sqrt{1-\alpha_{t-1}-\sigma_t^2}\epsilon_\theta(x_t,t)+\sigma_t^2z,
  \label{eq:8}
\end{equation}
where $f_\theta$ is the prediction of $x_0$ at $t$:
\begin{equation}
  f_\theta(x_t,t)=\frac{x_t-\sqrt{1-\alpha_t}\epsilon_\theta(x_t,t)}{\sqrt{\alpha_t}}.
  \label{eq:final}
\end{equation}
By setting $\sigma_t=0$ in \cref{eq:8}, the sampling process from $x_T$ to $x_0$ becomes deterministic, which is the principle of DDIM. Also, the operation of DDIM inversion can map $x_0$ back to $x_T$ by reversing the process, enabling subsequent editing of images. Deterministic DDIM sampling and inversion can be expressed as:
\begin{equation}
\begin{aligned}
    x_{t-1}&=\sqrt{\alpha_{t-1}}f_\theta(x_t,t)+\sqrt{1-\alpha_{t-1}}\epsilon_\theta(x_t,t),\\
    x_{t+1}&=\sqrt{\alpha_{t+1}}f_\theta(x_t,t)+\sqrt{1-\alpha_{t+1}}\epsilon_\theta(x_t,t).
\end{aligned}
  \label{eq:ddim}
\end{equation}
\begin{figure}
  \centering
    \includegraphics[width=\linewidth]{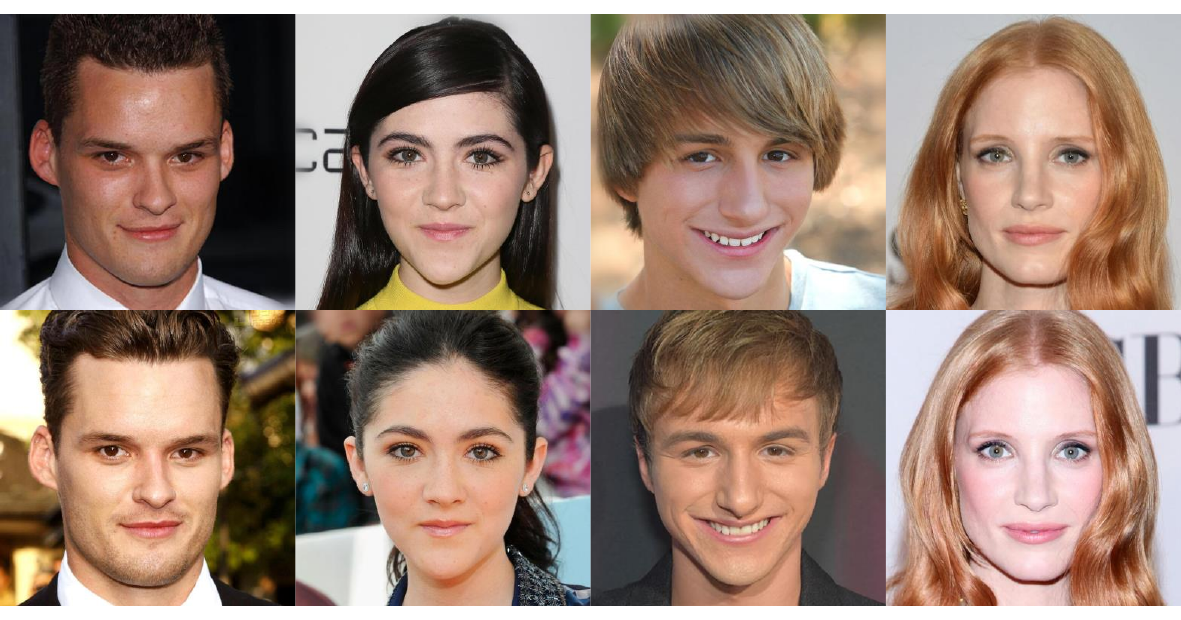}
    \caption{Target identities. The images in the top row are used for training, while the images in the bottom row are used for testing.}
    \label{fig:target}
\end{figure}
\begin{figure*}
  \centering
    \includegraphics[width=0.82\linewidth]{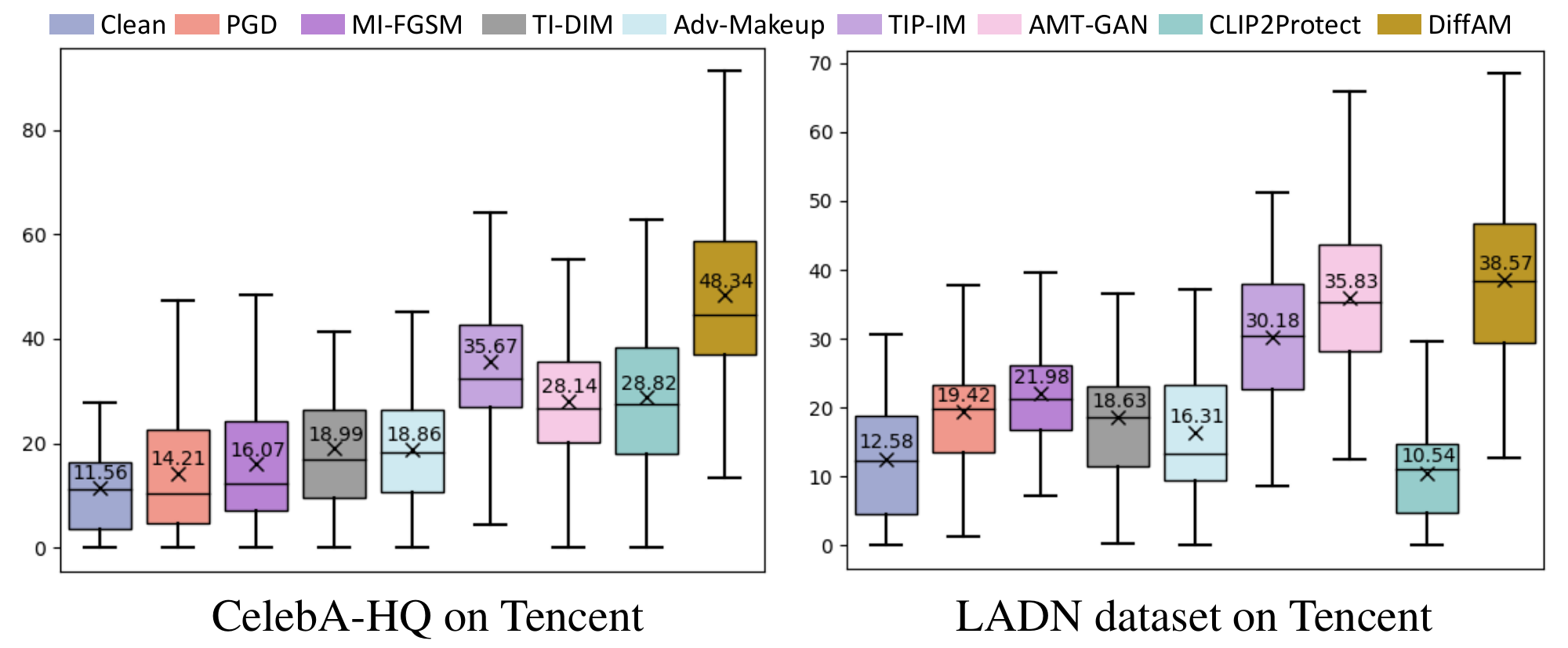}
    \caption{The confidence scores (higher is better) returned from Tencent API.}
    \label{fig:tencent}
\end{figure*}
\begin{figure*}[htbp]
\setlength{\belowcaptionskip}{-0.2cm}
  \centering
    \includegraphics[width=\linewidth]{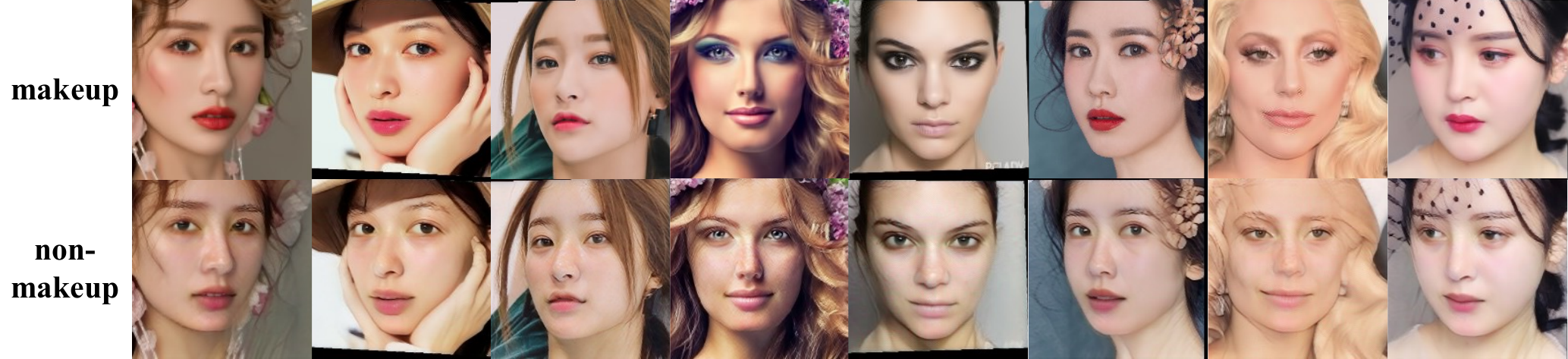}
    \caption{Visualizations of text-guided makeup removal. The top row shows some reference makeup images. The bottom row shows the corresponding non-makeup images generated by text-guided makeup removal module.}
    \label{fig:removal}
\end{figure*}
\begin{figure*}[htbp]
    \centering
    \begin{subfigure}{1\linewidth}
        \includegraphics[width=\textwidth]{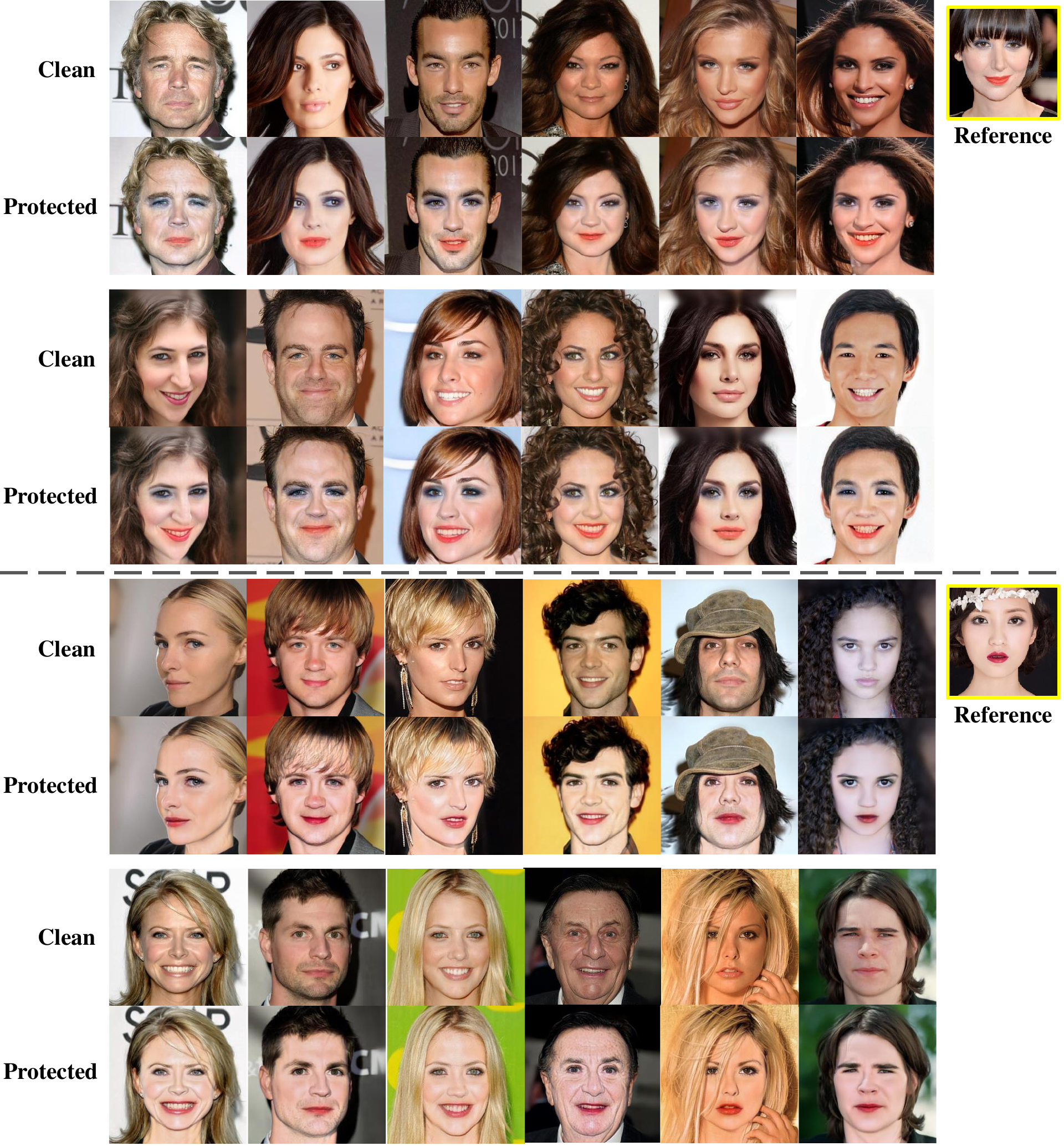}
    \end{subfigure}
\end{figure*}
\begin{figure*}[htbp]\ContinuedFloat
    \centering
    \begin{subfigure}{1\linewidth}
        \includegraphics[width=\textwidth]{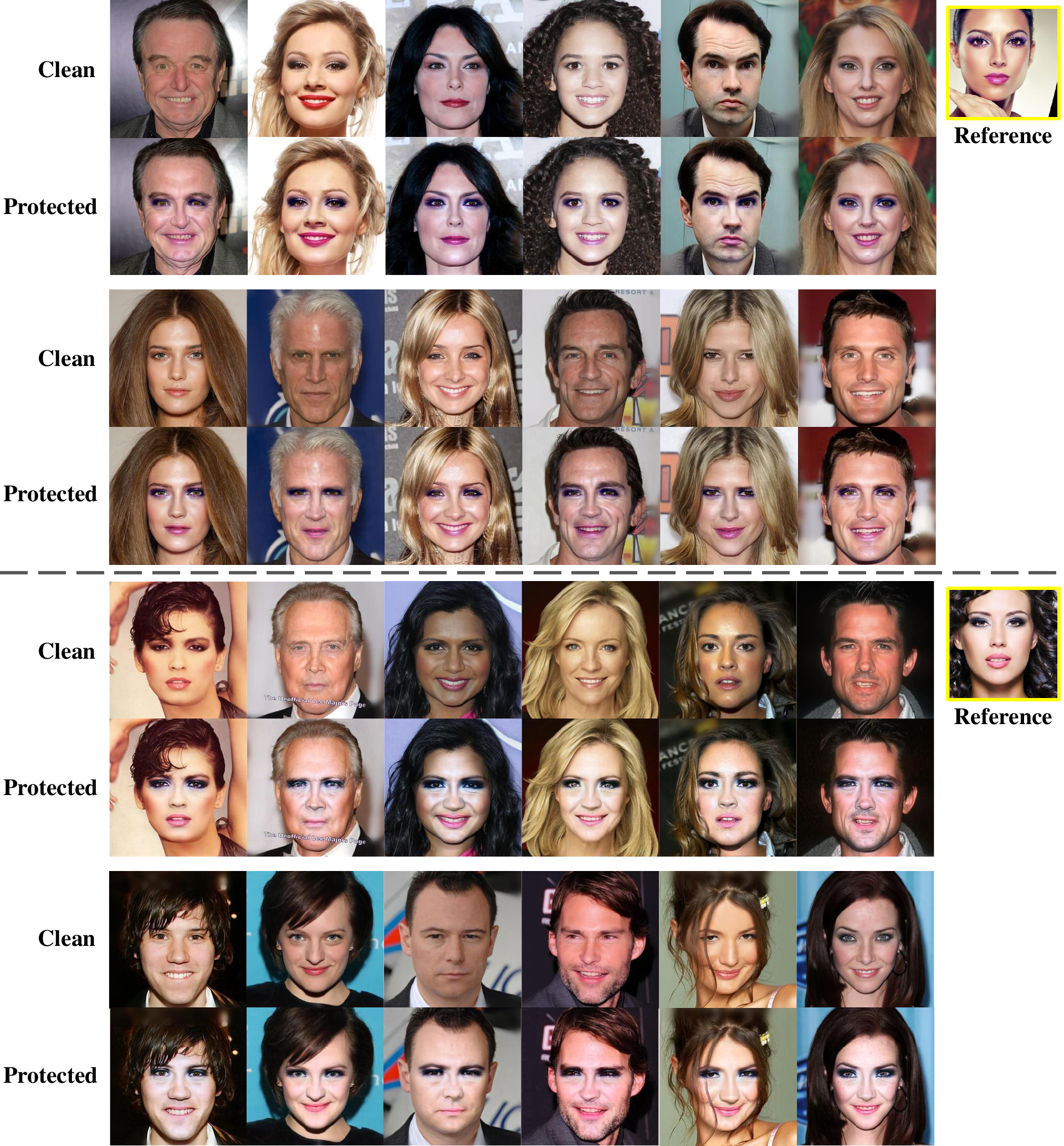}
    \end{subfigure}
\end{figure*}
\begin{figure*}[htbp]\ContinuedFloat
    \centering
    \begin{subfigure}{1\linewidth}
        \includegraphics[width=\textwidth]{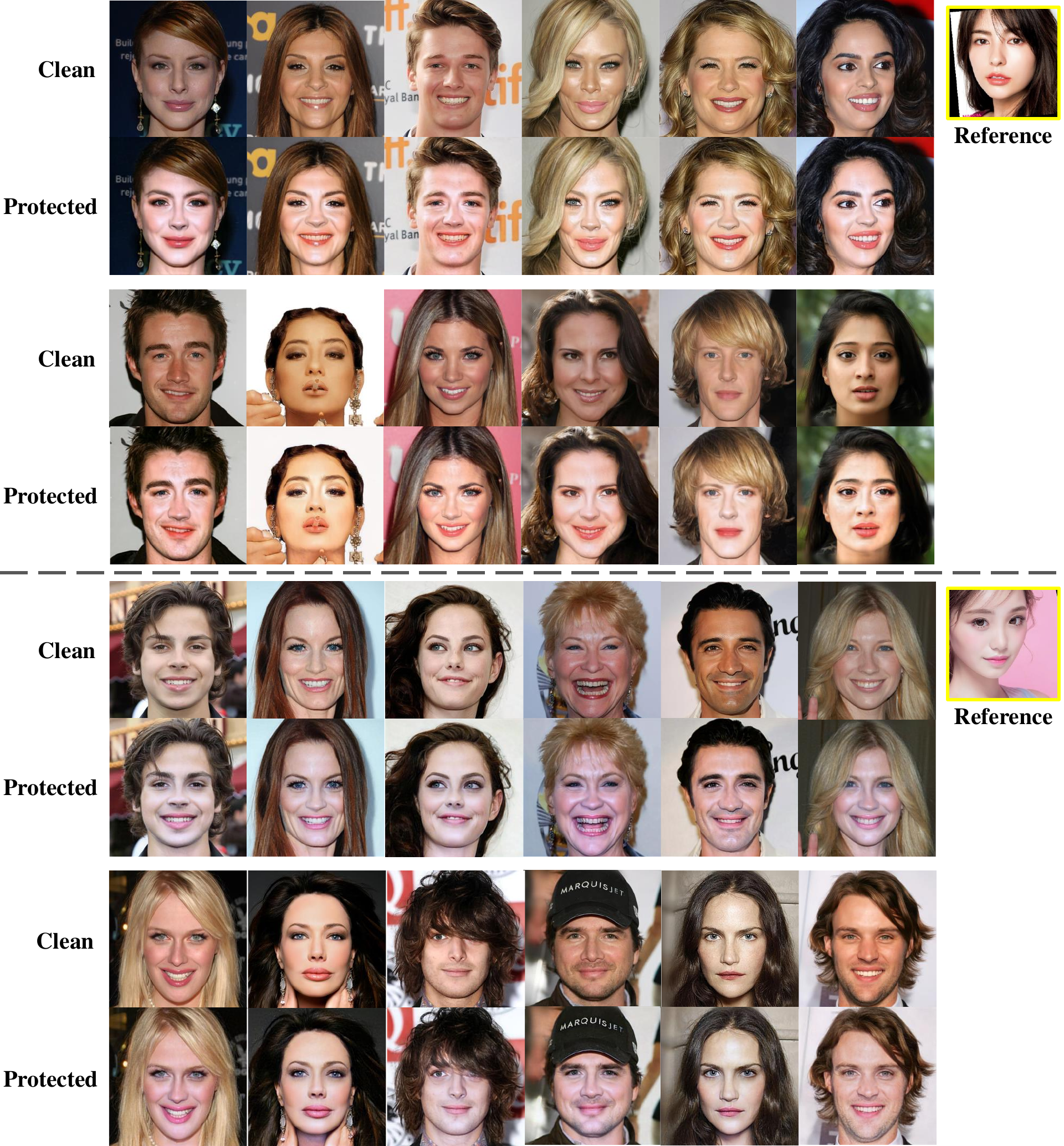}
    \end{subfigure}
    \caption{Visualizations of image-guided makeup transfer.}
    \label{fig:transfer}
\end{figure*}
\section{Target Images}
DiffAM aims to generate protected face images which mislead FR models into identifying them as the target identity. So we show the four target identities, provided by \cite{hu2022protecting}, used for our experiments in \cref{fig:target}. To simulate real-world attack scenarios, the target images used during training and testing are different. 

\section{Attack Performance on Tencent API}
To fully evaluate attack effects of DiffAM in real-world scenarios, we also shows the quantitative results of attacks on Tencent API\footnote{\url{https://cloud.tencent.com/product/facerecognition}} here in \cref{fig:tencent}. We randomly selected 100 images each from CelebA-HQ and LADN datasets to protect and report confidence scores returned from APIs. The confidence scores are between 0 to 100, where the higher score indicates higher similarity between the protected face image and the target image. DiffAM achieves the highest average confidence scores (\textbf{48.34} and \textbf{38.57}) campared to other methods. This further demonstrates that our precise guidance on adversarial makeup domain and robust adversarial makeup generation ensure high black-box transferability of protected face images generated by DiffAM.
\section{More Visual Results}
\noindent \textbf{Text-guided Makeup Removal} \cref{fig:removal} shows some visual results of text-guided makeup removal. The makeup of reference images, such as lipstick and eyeshadow, are clearly removed, indicating the powerful ability of text guidance in makeup removal. Then the difference in CLIP space between the makeup and non-makeup images can determine accurate makeup direction for subsequent makeup transfer.

\noindent \textbf{Image-guided Makeup Transfer} 
Due to space limitations of the main text, more visual results of image-guided makeup transfer are shown in \cref{fig:transfer}. DiffAM achieves precise makeup transfer for each given reference image and generates natural-looking protected face images, thanks to our precise control over makeup direction and distance.

\end{document}